\documentclass{article}

\usepackage{arxiv}

\usepackage{amsmath}
\usepackage[capposition=top]{floatrow}

\usepackage{amsmath}
\usepackage{algorithm}
\usepackage[noend]{algpseudocode}

\usepackage[utf8]{inputenc} % allow utf-8 input
\usepackage[T1]{fontenc}    % use 8-bit T1 fonts
\usepackage{hyperref}       % hyperlinks
\usepackage{url}            % simple URL typesetting
\usepackage{booktabs}       % professional-quality tables
\usepackage{amsfonts}       % blackboard math symbols
\usepackage{nicefrac}       % compact symbols for 1/2, etc.
\usepackage{microtype}      % microtypography
\usepackage{lipsum}		% Can be removed after putting your text content
\usepackage{graphicx}
\usepackage{natbib}
\usepackage{doi}
\usepackage{dsfont}
\usepackage{xcolor}

\newtheorem{remark}{Remark}[section]

\title{Risk Averse Non-Stationary Multi-Armed Bandits}

\author{ \href{}{\includegraphics[scale=0.06]{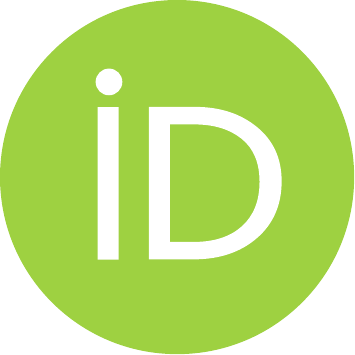}\hspace{1mm}Leo Benac} \\
	Department of Mathematics \& Statistics\\
	Concordia University\\
	Montreal, QC \\
	\texttt{benac.leo@gmail.com} \\
	\And
	\href{}{\includegraphics[scale=0.06]{orcid.pdf}\hspace{1mm}Frederic Godin\thanks{Corresponding author. Frédéric Godin gratefully acknowledges financial support from the Natural Sciences and Engineering Research Council of Canada (NSERC, grant number RGPIN-2017-06837).}} \\
	Department of Mathematics \& Statistics\\
	Concordia University\\
	Montreal, QC \\
	\texttt{frederic.godin@concordia.com} \\
}

\renewcommand{\shorttitle}{\textit{arXiv} Template}

\hypersetup{
pdftitle={A template for the arxiv style},
pdfsubject={q-bio.NC, q-bio.QM},
pdfauthor={Frederic Godin, Leo Benac},
pdfkeywords={First keyword, Second keyword, More},
}

\begin{document}

\maketitle
\begin{abstract}
	This paper tackles the risk averse multi-armed bandits problem when incurred losses are non-stationary. The conditional value-at-risk (CVaR) is used as the objective function. Two estimation methods are proposed for this objective function in the presence of non-stationary losses, one relying on a weighted empirical distribution of losses and another on the dual representation of the CVaR. Such estimates can then be embedded into classic arm selection methods such as $\epsilon$-greedy policies. Simulation experiments assess the performance of the arm selection algorithms based on the two novel estimation approaches, and such policies are shown to outperform naive benchmarks not taking non-stationarity into account.
\end{abstract}
% keywords can be removed

%\rc{I recommend avoiding hyperlinks to webpages; these webpages might not be active in a few years, whereas papers are expected to last forever.}

%%%%%%%%%%%%%%%%%%%%%%%%%%%%%%%%%%%%%%%%%%%%%%%%%%%%%%%%%%%%%%%%%%%%%%%%%%%%%%%%%%%%%
%%%%%%%%%%%%%%%%%%%%%%%%%%%%%%%%%%%%%%%%%%%%%%%%%%%%%%%%%%%%%%%%%%%%%%%%%%%%%%%%%%%%%

\section{Introduction}
In a stochastic multi-armed bandits (MAB) problem, an agent is repeatedly faced with the choice of sampling from one of $K$ arms, each providing rewards drawn from an unknown distribution. The agent seeks to find proper balance between exploration and exploitation so as to optimize its objective related to rewards collected during a set of trials. Such a framework has multiple applications such as recommendation systems, advertising, finance and medical trials among others, see \cite{bouneffouf2019survey}. Classic versions of the problem are concerned with maximizing expected rewards, as in \cite{sutton1998introduction}. More recent work substitute the expected reward maximisation objective with risk minimization. This is motivated by situations were an adverse outcome on any draw can have very detrimental consequences. Such risk averse MAB problems have applications in health science (e.g. medical trials), robotics and energy management, see for instance \cite{galichet2013exploration}. Risk aversion metrics considered in the literature include for instance a mean-variance trade-off criterion \citep{sani2012risk}, or tail risk measures such as in \cite{yu2013sample}. A popular example of a tail risk measure used in the context of risk averse MAB problems is the conditional value-at-risk (CVaR) introduced by \cite{rockafellar2002conditional}, which quantifies the expected loss (i.e. minus the reward) given it exceeds a given quantile of its distribution. 
The present work therefore also considers the same risk measure, despite other potential choices being available such as the value-at-risk (VaR), or expectiles \citep{newey1987asymmetric}. Other risk-averse MAB papers also considered the CVaR. Upper confidence bound algorithms in this context are studied by \cite{maillard2013robust}, \cite{cassel2018general}, \cite{khajonchotpanya2021revised}. Alternative arm selection approaches in the context of risk-averse bandits include the max-min approach discussed in \cite{galichet2013exploration}, the successive rejects relying on concentration bound guarantees of \cite{kolla2019risk}, robust estimation-based algorithms in \cite{kagrecha2020statistically}, or Thompson Sampling approaches in \cite{chang2020risk} and \cite{baudry2021optimal}.

Papers from the risk averse MAB stream mainly consider stationary problems where the loss distribution for each arm remains the same over all trials. In practice, several real-life problems involve cases where the loss distributions progressively fluctuate through time.
%(For the former case, some stocks that changes values everyday could be a real life example of the distributions changing progressively. In medical context however we can have medicines that works very well depending on the season we take it. A medicine taken in winter could have a total different effect that one taking in summer for example.)
Such changes in distributions are referred to as non-stationarities. The present work thus fills a gap from the literature by being, to the best of authors' knowledge, the first to consider non-stationary risk averse multi-armed bandits problems. To tackle such problems, two estimation procedures for the CVaR in the presence of non-stationary losses is proposed. The first one relies on a weighted empirical distribution of losses. The second couples the dual representation of the CVaR with the traditional move-toward-target update formula to estimate expected auxiliary functions found within the latter representation.
Numerical experiments show that in a non-stationary losses context, the proposed CVaR estimation methods exhibit significant outperformance over a more naive sample averaging approach not considering the evolution of distributions. 

The paper is divided as follows. Section \ref{se:CVaRestim} describes the CVaR risk measure and outlines how related estimates of risk can be obtained. The risk averse multi-armed bandits setting is discussed in Section \ref{se:MAB}. Results from numerical experiments assessing the performance of proposed methods for a non-stationary risk averse multi-armed bandits problem are provided in Section \ref{se:NumExp}. Section \ref{se:Conclusion} concludes.

\section{CVaR risk measure and estimation}
\label{se:CVaRestim}

The CVaR measure originally proposed by \cite{rockafellar2002conditional} is meant to measure the average loss in a set of worst-case scenarios, thereby reflecting the severity associated with extreme losses.
For a loss random variable $Z$, denote its cumulative distribution function (cdf) by $F_Z$. The CVaR associated with a confidence level $\alpha$ can be formally defined as
\begin{eqnarray*}
    \text{CVaR}_\alpha(Z) &\equiv& \frac{1}{1-\alpha} \int^1_{\alpha} q_u(Z) du, \quad \text{where}
    \\ q_\alpha(Z) &\equiv& \text{inf}\{z \in R: F_{Z}(z) \geq \alpha \}.
\end{eqnarray*}
Typical values for $\alpha$ include $0.9$, $0.95$ and $0.99$. $q_\alpha(Z)$ is the quantile of confidence level $\alpha$ of the distribution $Z$. If $Z$ is an absolutely continuous random variable (i.e. it is atomless), then the CVaR possesses the following intuitive representation justifying its interpretation as the expected loss in worst-case scenarios:
\begin{equation*}
    \text{CVaR}_\alpha(Z) = E[Z | Z \geq q_\alpha(Z)].
\end{equation*}
Moreover, \cite{rockafellar2002conditional} provide the following dual representation for the CVaR which will subsequently be handy in the present work:
\begin{eqnarray}
\label{dualrep}
\text{CVaR}_{\alpha}(Z) &=& \min\limits_{c \in\mathbb{R}}\mathbb{E}  [f^{\text{CVaR}}_{c,\alpha}(Z) ], \quad \text{where}
\\ f^{\text{CVaR}}_{c,\alpha}(z) &\equiv& c + \frac{1}{1- \alpha} (z - c) \mathds{1}_{\{z > c\}}. \notag
\end{eqnarray}
Such dual representation expresses the CVaR through a set of unconditional expectations, which can be evaluated conveniently with typical statistical and reinforcement learning techniques. Furthermore, it is shown in \cite{rockafellar2002conditional} that
\begin{equation*}
    \arg\!\min\limits_{c \in\mathbb{R}}\mathbb{E}  [f^{\text{CVaR}}_{c,\alpha}(Z) ] = q_\alpha(Z),
\end{equation*}
which means that the minimizing auxiliary constant $c$ is the quantile of level $\alpha$ of the distribution of $Z$.

\subsection{Estimation with an i.i.d sample}

A required endeavour in a multi-armed bandits framework is the estimation of the objective function (the CVaR in the present case) applied to the a variable $Z$ through a sample $Z_1,\ldots,Z_n$. When the sample contains independent and identically distributed (i.i.d.) observations, a straightforward approach consists in using the empirical distribution of the sample as an approximation of the true distribution of $Z$. For the CVaR risk measure, this leads to the following sample averaging formula \citep[see for instance][]{troop2021bias}:
\begin{equation}
  \label{eq:TCEestimsample}
  \widehat{\text{CVaR}}^{(avg)}_{\alpha,n}(Z)  
  = \frac{\sum_{i=1}^{n} Z_{i} \mathds{1}_{ \{Z_i \geq \hat{q}^{n}_{\alpha}(Z) \} }}{\sum_{i=1}^{n} \mathds{1}_{ \{Z_i \geq \hat{q}^{n}_{\alpha}(Z) \} }}.
\end{equation}
where the empirical cdf of $Z$ is given by
\begin{equation}
  \label{SampleCDFestim}
  \hat{F}^{n}_Z(z) \equiv n^{-1} \sum_{s=1}^{n} \mathds{1}_{ \{ Z_s \leq z \} },
\end{equation}
and the empirical distribution quantiles are provided by
\begin{eqnarray*}
  %\label{quantileEstim}
  \hat{q}^{n}_{\alpha}(Z) \equiv \inf\{z\in\mathbb{R} : \hat{F}^{n}_Z(z) \geq \alpha\} = \underset{i }{\min}\{Z_{(i)} : \hat{F}^{n}_Z(Z_{(i)}) \geq \alpha\} = Z_{(\lceil\alpha n\rceil)},
\end{eqnarray*}
with $Z_{(1)},\ldots, Z_{(n)}$ being the order statistics (i.e. sample observations sorted in a non-decreasing order). \cite{trindade2007financial} study among others the asymptotic behavior of such an empirical distribution-based estimator, and thereby show its consistency. This CVaR estimation approach is referred to as the \textit{sample averaging} method.

\subsection{Estimation under non-stationarities}

When the distributions underlying observations $Z_1,\ldots,Z_n$ are not identical, there is no single CVaR to estimate. In a non-startionary MAB setting, the required task would consist in estimating $\text{CVaR}_\alpha(Z_{n+1})$ based on the available sample, under the assumption that the fluctuation in the distribution of $Z_t$ between any time $t$ and $t+1$ is not too severe.

The main idea considered in the present study, which is also illustrated in \cite{sutton1998introduction}, consists in putting more weight on more recent observations, since the distribution of $Z_{n+1}$ is most likely more similar to that of recent observations $Z_{n}, Z_{n-1}, \cdots,$ rather than to that of observations further away (e.g. $Z_1,Z_2,\ldots$). Therefore, consider weights $w^{(n+1)}_1,\ldots,w^{(n+1)}_n$ associated with observations $Z_1,\ldots,Z_n$ when predicting the CVaR for observation $Z_{n+1}$. Exponential decay could be for instance considered to assign more weight to more recent observations. For some $\lambda>0$, this can be accomplished by setting
\begin{equation*}
    w^{(n+1)}_j = (1-\lambda)^{n-j} \frac{1-(1-\lambda)}{1-(1-\lambda)^n}, \quad j=1,\ldots,n,
\end{equation*}
which satisfies $\sum^n_{j=1} w^{(n+1)}_j =1$ and $w^{(n+1)}_{j} = (1-\lambda) w^{(n+1)}_{j+1}, \, j=1,\ldots,n-1$. Therefore, a value of $\lambda$ close to zero leads to close to uniform weights across all observations, whereas a $\lambda$ close to one puts the majority of the weight on very recent observations.

This leads to a first potential estimator based on a weighted empirical distribution of losses:
\begin{eqnarray}
     \widehat{\text{CVaR}}^{(weight)}_{\alpha,n}(Z_{n+1})  
  &\equiv& \frac{\sum_{i=1}^{n} w^{(n+1)}_i Z_{i} \mathds{1}_{ \{Z_i \geq \tilde{q}^{n}_{\alpha}(Z_{n+1}) \} }}{\sum_{i=1}^{n} w^{(n+1)}_i \mathds{1}_{ \{Z_i \geq \tilde{q}^{n}_{\alpha}(Z_{n+1}) \} }}, \quad \text{where} \label{weightedCVaR}
  \\  \tilde{F}^{n}_{Z_{n+1}}(z) &\equiv&\sum_{s=1}^{n} w^{(n+1)}_{s}\mathds{1}_{ \{ Z_s \leq z \} }, \notag
  \\  \tilde{q}^{n}_{\alpha}(Z_{n+1}) &\equiv& \inf\{z\in\mathbb{R} : \tilde{F}^{n}_{Z_{n+1}}(z) \geq \alpha\}. \notag
\end{eqnarray}
This method is referred to as the \textit{weighted empirical estimation} approach.
%\rc{I must remove the Z next to chat + arrange notation, I'll do that tomorrow}

An issue with that approach is that the stage-$t$ quantile estimate $\hat{q}^{t}_{\alpha}(Z_{t+1})$ fluctuates across stages $t$. Thus \eqref{weightedCVaR} cannot lead to a simple recursive formula linking $\widehat{\text{CVaR}}^{(weight)}_{\alpha,n}(Z_{n+1})$ and $\widehat{\text{CVaR}}^{(weight)}_{\alpha,n-1}(Z_n)$.
%\rc{Perhaps we should nevertheless try the above estimator in the numerical experiments.}
It would be convenient to include exponential weighting within the following recursive approach outlined in \cite{sutton1998introduction}:
\begin{equation}
\label{recursGen}
    \text{New estimate} = \text{Past estimate} + \text{Learning rate} \times ( \text{Target} - \text{Past estimate}).
\end{equation}
Advantages of using an algorithm along the lines of \eqref{recursGen} are that (i) it works well in an on-line fashion, e.g. if updates need to be run in parallel and/or in real-time, and (ii) it doesn't require storing the entire history of losses incurred unlike \eqref{weightedCVaR}.
Such recursive update formula works conveniently when the objective function to estimate is an expected value. This is where the dual representation of the CVaR becomes useful. Indeed, denote by $\mathcal{E}_{n,c,\alpha}$ the estimate of the quantity $\mathbb{E}  [f^{\text{CVaR}}_{c,\alpha}(Z_{n+1}) ]$ found in \eqref{dualrep}. Then, because $\mathcal{E}_{n,c,\alpha}$ is an expectation, a recursion of the type \eqref{recursGen} can be obtained through 
\begin{equation}
\label{updateDual}
    \mathcal{E}_{n+1,c,\alpha} \equiv \mathcal{E}_{n,c,\alpha} + \lambda \left(f^{CVaR}_{c,\alpha}(Z_n) - \mathcal{E}_{n,c,\alpha}\right), \quad n=1,2,\ldots
\end{equation}
for some $\lambda >0$ with given starting values $\mathcal{E}_{1,c,\alpha}$, which leads to the following estimate of the $\text{CVaR}_\alpha$ due to \eqref{dualrep}:
\begin{equation}
\label{CVaRestimNonstat}
    \widehat{\text{CVaR}}^{(recurs)}_{\alpha,n}(Z_{n+1}) \equiv \underset{c \in \mathbb{R}}{\min} \, \mathcal{E}_{n,c,\alpha}.
\end{equation}
Such an estimation approach is referred to as the \textit{dual recursive estimation} method. Similarly to the weighted empirical estimation approach, the higher the value for $\lambda$, the more impact recent observations have on estimates relatively to earlier observations.  Estimates \eqref{weightedCVaR} and \eqref{CVaRestimNonstat} are not identical; this can be seen for instance with $\mathcal{E}_{n,c,\alpha}$, $n>0$ depending on starting values $\mathcal{E}_{1,c,\alpha}$, unlike $\widehat{\text{CVaR}}^{(weight)}_{\alpha,n}(Z_{n+1}) $ in \eqref{weightedCVaR}. 

%%%%%%%%%%%%%%%%%%%%%%%%%%%%%%%%%%%%%%%%%%%%%%%%%%%%%%%%%%%%%%%%%%%%
%%%%%%%%%%%%%%%%%%%%%%%%%%%%%%%%%%%%%%%%%%%%%%%%%%%%%%%%%%%%%%%%%%%%

\section{Multi-armed bandits setting \label{se:MAB}}

In the multi-armed bandits setting, there are $K$ arms which can be sampled, and the loss\footnote{Bandits problems are often expressed in terms of rewards. However due to the risk averse nature of the agent considered herein, we shall refer to losses which could be understood as minus the rewards.} provided by arm $i$, at stage $t$ (if it is sampled) is denoted $Y_{t,i}$, $t=1,\ldots,T$. Denote by $F_{Y_{t,i}}$ the cdf of such reward. Different goals can be pursued in such context. In pure exploration problems, the goal is simply to attempt identifying the least risky arm (i.e. the one with the smallest CVaR$_\alpha$ with $\alpha$ being given in the problem) within $T$ trials. In other problems, the performance associated with losses incurred over the $T$ stages is important and the objective is to reduce the amount of extreme losses incurred during the run, where the run is defined as the action of going through all time stages and incurring associated losses.
To achieve the desired goal, the agent must select on each stage $t$ an arm (i.e. an action), which is denoted by $a_t$. The set of all actions is characterized by a \textit{policy} which maps available information into an action. Thus a policy $\pi = \{\pi_t\}^T_{t=1}$ contains the sequence of mappings $\pi_t : (a_1,Y_{1,a_1},\ldots,a_{t-1},Y_{t-1,a_{t-1}}) \rightarrow p_t$, $t=1,\ldots,T$, where $p_t$ is the random vector containing probabilities of selecting any action in $\{1,\ldots,K\}$ at stage $t$.
%for each realization of the probability space $\omega \in \Omega$ the action corresponds to an arm i.e. $a_t(\omega) \in \{1,\ldots,K\}$.

For simplicity and due to their popularity, $\epsilon$-greedy policies are considered in the present study. These entail that in a given stage $t$, the greedy action (i.e. the one with the smallest estimated CVaR) is selected with probability $1-\epsilon$, and otherwise with probability $\epsilon$ an action is randomly sampled across all arms (including the greedy one).\footnote{ Due to the possibility of sampling the greedy action when exploring, the true probability of selecting the greedy action is in fact $1-\epsilon + \epsilon/K$. } Defining an i.i.d. sequence $\{H_t\}^T_{t=1}$ of Bernoulli$(\epsilon)$ random variables where $H_t$ is independent from previous losses and actions $\left(a_1,Y_{1,a_1}\right),\ldots,\left(a_{t-1},Y_{t-1,a_{t-1}}\right)$, actions underlying the $\epsilon$-greedy policy can be represented as
\begin{equation*}
    a_t = H_t A_t+ (1-H_t) \hat{a}^*_t
\end{equation*}
where $A_t$ is uniformly sampled among $\{1,\ldots,K\}$ independently of previous realized actions and losses, and $\hat{a}^*_t$ is the greedy actions defined as
\begin{equation}
\label{greedyaction}
    \hat{a}^*_t \equiv \underset{i=1,\ldots,K}{\arg \! \min} \,  \widehat{\text{CVaR}}_{\alpha,t-1}(Y_{t,i}).
\end{equation}
Each estimation method for the CVaR therefore leads to a different estimate of the greedy action. Larger values of $\epsilon$ are associated with more exploration, forcing the agent to try non-greedy actions more often to refine knowledge about their distributions, whereas a smaller $\epsilon$ entails more exploitation by generating losses from actions deemed less risky based  on current estimates. In the presence of non-stationary losses, exploration is even more important as older estimates eventually become obsolete due to changes in the loss distributions. Note that to form the estimate $\widehat{\text{CVaR}}_{\alpha,t-1}(Y_{t,i})$ in \eqref{greedyaction}, the estimator (either \eqref{eq:TCEestimsample}, \eqref{weightedCVaR} or \eqref{CVaRestimNonstat}) is applied on the set of observed losses provided by arm $i$: $\mathcal{S}^t_{i} \equiv \{ Y_{u,i} : a_u = i, \, u=1,\ldots,t-1 \}$.% noting that \eqref{CVaRestimNonstat} in fact only relies on the most recent observation in the sample $\mathcal{S}^t_{i}$ to perform recursive estimation \eqref{updateDual}.

%\rc{Reword to explain the last one does not rely on the entire sample to fully re-estimate each step}

The objective in subsequent numerical experiments consists in empirically detailing the performance of the various estimation methods (including the choice of parameter $\lambda$) when embedded in the $\epsilon$-greedy policy. The first strategy considered, which acts as a benchmark, is the application of the sample averaging method \eqref{eq:TCEestimsample}. Algorithm \ref{euclid} provides the pseudo-code for the $\epsilon$-greedy policy under such a method to estimate the CVaR for all arms.

\begin{algorithm}
\caption{Sample averaging estimation algorithm}\label{euclid}
\begin{algorithmic}[1]
%\Procedure{MyProcedure}{}
\vspace{0.05 cm}
\State \emph{\textbf{Inputs}}: $\alpha\in(0,1)$, $\lambda>0$, $\epsilon \in (0,1)$
\vspace{0.05 cm}
\State \emph{\textbf{Loop} over all stages $t$}
\vspace{0.05 cm}
\State \hspace{0.5 cm} Sample $H_t \sim$ Bernoulli$(\epsilon)$
\vspace{0.05 cm}
\State \hspace{0.5 cm} \textbf{If} $H_t =0$
\vspace{0.05 cm}
\State \hspace{1 cm} $a_t \gets \underset{i=1,\ldots,K}{\arg \!\min}  \hspace{0.1 cm} \widehat{\text{CVaR}}^{(avg)}_{\alpha,t-1}(Y_{t,i})$
\vspace{0.05 cm}
\State \hspace{0.5 cm} \textbf{Else}
\vspace{0.05 cm}
\State \hspace{1 cm} Sample $a_t$ uniformly from $\{1,\ldots,K \}$
%\vspace{0.05 cm}
%\State \hspace{0.5 cm} \textbf{End if}
\vspace{0.05 cm}
\State \hspace{0.5 cm} Observe $Y_{t,a_t}$, the time-$t$ reward from arm $a_t$
\vspace{0.05 cm}
\State \hspace{0.5 cm} Calculate $\widehat{\text{CVaR}}^{(avg)}_{\alpha,t}(Y_{t+1, a_t})$ by applying \eqref{eq:TCEestimsample} on sample $\mathcal{S}_{a_t}^t$
\vspace{0.05 cm}
\State \hspace{0.5 cm} $\widehat{\text{CVaR}}^{(avg)}_{\alpha,t}(Y_{t+1, i}) \gets \widehat{\text{CVaR}}^{(avg)}_{\alpha,t-1}(Y_{t, i})$ for all arms $i \neq a_t$
\end{algorithmic}
\end{algorithm}

For the weighted empirical estimation approach, the algorithm considered is the exact same as Algorithm \ref{euclid}, except that \eqref{weightedCVaR} is applied on line 9 instead.
%\end{remark}

The performance of the proposed estimation strategy \eqref{updateDual}-\eqref{CVaRestimNonstat} allowing to consider non-stationarities is also assessed. In theory, \eqref{CVaRestimNonstat} requires having the estimate of $\mathcal{E}^{(i)}_{t,c,\alpha}$ for all $c \in \mathbb{R}$ and for all arms $i$ (as each arm must have distinct values in a MAB setting, hence the added superscript). In practice, estimates can only be stored for a finite number of values for $c$. Therefore, a grid $\mathbb{G}\equiv \{c_1,\ldots,c_M \}$ containing all values that are considered for $c$ is introduced, and the following approximation is applied during the implementation of the method:
\begin{equation*}
    \underset{c \in \mathbb{R}}{\min} \, \mathcal{E}^{(i)}_{t,c,\alpha} \approx \underset{c \in \mathbb{G}}{\min} \, \mathcal{E}^{(i)}_{t,c,\alpha},
\end{equation*}
which should be valid provided that the grid $\mathbb{G}$ is sufficiently fine and its range is sufficiently large.  

Algorithm \ref{euclid2} outlines the pseudo-code for the application of the $\epsilon$-greedy policy used in conjunction with estimation method \eqref{updateDual}-\eqref{CVaRestimNonstat}. The approach considered involves refining values of $\mathcal{E}^{(i)}_{t,c,\alpha}$ for all $c \in \mathbb{G}$ whenever an action $i$ is sampled.

%\subsection{Non-stationary Algorithm}

%We just stated that we can express $\textit{CVaR}_{\alpha}(Y_{t, a^t})$ in terms of $\mathbb{E}[Y_{t, a^t}]$. In the non-stationary context, we can refine our estimates as follows:

%\hspace{5cm}$\hat{\mathbb{E}}[f_{c}^{ CVaR_{\alpha}^{(t)}  }(Y_i)] = {\hat{CVaR}_{\alpha}^{t}(Y)} $
%\hspace{4cm} $\textit{target}_c = c\hspace{3.32 cm}  \textit{if } r^t \le c$
%\hspace{4 cm} $ \textit{target}_c = c + \frac{1}{1 - \alpha}(r^t - c) \hspace{1.1 cm} \textit{if } r^t > c$
%\hspace{4cm} $\hat{\mathbb{E}}[f_{c}^{ (\textit{CVaR}_{\alpha} ) }(Y_{t+1, a^t})] = \hat{\mathbb{E}}[f_{c}^{ (\textit{CVaR}_{\alpha} ) }(Y_{t,a^t})] + \lambda * (target_c - \hat{\mathbb{E}}[f_{c}^{ (\textit{CVaR}_{\alpha} ) }(Y_{t, a^t})] )  $
%\hspace{4cm}$\widehat{\textit{CVaR}}_{\alpha}(Y_{t+1, a^t}) = \min\limits_{c \in\mathbb{G}}\hat{\mathbb{E}}[f_{}^{ (\textit{CVaR}_{\alpha} )}(Y_{t+1, a^t})] $ \hspace{ 2 cm} (5)

%$\lambda$ is the stepsize and $\mathbb{G}$ is a grid of c values where we optimize our estimates as we cannot optimize over $\mathbb{R}$ directly. 
%For a fixed c value and time t, each arm has a corresponding estimate $\hat{\mathbb{E}}[f_{c}^{ (\textit{CVaR}_{\alpha} ) }(Y_{t,a^t})]$ that gets refine every time we sample from it. The same way in the traditional bandits framework we pick the initial value for the means of our arms, we can pick the values for each initial estimates $\hat{\mathbb{E}}[f_{c}^{ (\textit{CVaR}_{\alpha} ) }(Y_{1, i})]$.

\begin{algorithm}
\caption{Dual recursive estimation algorithm}\label{euclid2}
\begin{algorithmic}[1]
%\Procedure{MyProcedure}{}
\vspace{0.05 cm}
\State \emph{\textbf{Inputs}}: $\alpha\in(0,1)$, $\lambda>0$, $\epsilon \in (0,1)$, grid $\mathbb{G}$, initial estimates $\mathcal{E}_{1,c,\alpha}$ for all $c \in \mathbb{G}$
\vspace{0.05 cm}
\State \emph{\textbf{Loop} over all stages $t$}:
\State \hspace{0.5 cm} Sample $H_t \sim$ Bernoulli$(\epsilon)$
\vspace{0.05 cm}
\State \hspace{0.5 cm} \textbf{If} $H_t =0$
\vspace{0.05 cm}
\State \hspace{1 cm} $a_t \gets \underset{i=1,\ldots,K}{\arg \!\min}  \, \underset{c \in \mathbb{G}}{\min} \, \mathcal{E}^{(i)}_{t,c,\alpha}$
\vspace{0.05 cm}
\State \hspace{0.5 cm} \textbf{Else}
\vspace{0.05 cm}
\State \hspace{1 cm} Sample $a_t$ uniformly from $\{1,\ldots,K \}$
%\vspace{0.05 cm}
%\State \hspace{0.5 cm} \textbf{End if}
\vspace{0.05 cm}
\State \hspace{0.5 cm} Observe $Y_{t,a_t}$, the time-$t$ reward from arm $a_t$
%\State \hspace{0.5 cm} $\widehat{\textit{CVaR}}_{\alpha}(Y_{t, i}) \gets \min\limits_{c \in \mathbb{G}} \hat{\mathbb{E}}[f_{c}^{ (\textit{CVaR}_{\alpha} ) }(Y_{t,i})] \hspace{0.2 cm} \forall i  $
%\State \hspace{0.5 cm} $a^t \gets \underset{i}{argmin}  \hspace{0.1 cm} \widehat{\textit{CVaR}}_{\alpha}(Y_{t, i})$
%\State \hspace{0.5 cm} $r^t \gets \textit{pull arm}(a^t)$
\vspace{0.05 cm}
\State \hspace{0.5 cm} \emph{\textbf{Loop}} over all arms $i\neq a_t$ and all $c \in \mathbb{G}$, 
\vspace{0.05 cm}
\State \hspace{1 cm} $\mathcal{E}^{(i)}_{t+1,c,\alpha} \gets \mathcal{E}^{(i)}_{t,c,\alpha}$
\vspace{0.2 cm}
\State \hspace{0.5 cm} \emph{{\textbf{Loop}} over all c $\in \mathbb{G}$}:
\State \hspace{1 cm}$\textit{Target}_c \gets c + \frac{1}{1- \alpha} (Y_{t,a_t} - c) \mathds{1}_{\{Y_{t,a_t} > c\}}$
%\State \hspace{.9 cm} past estimate $ \gets \hat{\mathbb{E}}[f_{c}^{ (\textit{CVaR}_{\alpha} ) }(Y_{t, a^t})] $
%\State \hspace{.9 cm} $\hat{\mathbb{E}}[f_{c}^{ (\textit{CVaR}_{\alpha} ) }(Y_{t+1, a^t})] \gets $ past estimate + $\lambda * (\textit{target}_c$ - past estimate) 
\vspace{0.05 cm}
\State \hspace{1 cm} $\mathcal{E}^{(a_t)}_{t+1,c,\alpha} \gets \mathcal{E}^{(a_t)}_{t,c,\alpha} + \lambda \left(Target_c - \mathcal{E}^{(a_t)}_{t,c,\alpha} \right)$
\end{algorithmic}
\end{algorithm}

\begin{remark}
An important concern extensively studied in the MAB literature is the identification of performance guarantees, for instance through concentration bounds. Concentration bounds on CVaR estimates and related results under various assumptions are provided among others in \cite{brown2007large}, \cite{wang2010deviation}, \cite{kolla2019concentration} and \cite{prashanth2020concentration}. However, given the non-stationary nature of losses considered in the present work, the derivation of general concentration bounds coping with any possible form of non-stationarity is impossible, explaining why a simple exploratory simulation experiment rather than full-blown mathematical derivations are applied subsequently for performance analysis.
\end{remark}

%%%%%%%%%%%%%%%%%%%%%%%%%%%%%%%%%%%%%%%%%%%%%%%%%%%%%%%%%%%%%%%%%%%%%%%%%%%%%%%%%%%%%%
%%%%%%%%%%%%%%%%%%%%%%%%%%%%%%%%%%%%%%%%%%%%%%%%%%%%%%%%%%%%%%%%%%%%%%%%%%%%%%%%%%%%%%

\section{Numerical Experiments}
\label{se:NumExp}

This section details numerical experiments conducted to assess the performance of aforementioned policies in the context of risk averse non-stationary multi-armed bandits problem. The experiments take the form of Monte Carlo simulations and are exploratory (i.e. limited scope) in nature.

%%%%%%%%%%%%%%%%%%%%%%%%%%%%%%%%%%%%%%%%%%%%%%%%%%%%%%%%%%%%%%%%%%%%%%%%%%%%%%%%%%%%%%
%%%%%%%%%%%%%%%%%%%%%%%%%%%%%%%%%%%%%%%%%%%%%%%%%%%%%%%%%%%%%%%%%%%%%%%%%%%%%%%%%%%%%%

\subsection{Multi-armed testbed setting}

Experiments conducted in this section are analogous to the multi-armed testbed simulation performed in Chapter 2 of \cite{sutton1998introduction}. Several runs of the multi-arm bandits problem are simulated with the various competing policies. Performance metrics for each policy are calculated for each separate run, and are then aggregated across the runs to obtain the performance assessment.

In the experiment, $1,\!000$ runs are produced, with each run containing $2,\!000$ stages (i.e. time steps). $K=8$ arms are considered. The confidence level of the CVaR is set to $\alpha=0.90$. The exploration rate $\epsilon=0.05$ is used in $\epsilon$-greedy policies.

For the dual recursive estimation approach \eqref{updateDual}-\eqref{CVaRestimNonstat}, initial estimates of expected values of auxiliary losses are set to $\mathcal{E}^{(i)}_{1,c,\alpha}=0$ for all arms $i=1,\ldots,K$ and all $c$ in the grid $\mathbb{G}$. The grid $\mathbb{G}$ used in experiments 
%is of the form $\mathbb{G}=\{ (j-1)\frac{350}{999}: j=1,\ldots 1,000\}$, which 
contains $2,\!000$ equally spaced points ranging from $-100$ to $350$.
The following learning rates/weight decay values for $\lambda$  are tested: $0.01$, $0.05$, $0.1$, $0.2$, $0.3$, $0.4$, $0.5$, $0.6$, $0.7$, $0.8$, $0.9$, $0.95$, $0.99$.

Note that the true initial CVaR for each of the arms is always slightly higher than 0 for loss distributions considered in the experiment as detailed subsequently. Hence, setting $\mathcal{E}^{(i)}_{1,c,\alpha}=0$ for all arms $i=1,\ldots,K$ leads to optimistic initial values, which tends to force some exploration in early steps, see \cite{sutton1998introduction}.

%\rc{REVIEW VALUES ABOVE. The number of runs should be eventually increased (in progress)}

%, with two cases being considered. A first where distributions slowly evolve over time, and another with seldom abrupt shocks to distributions. This allows assessing how the proposed estimation method behaves under different types of non-stationarities which can be encountered in practice.

The specification of the evolution of loss distributions across time is provided subsequently. The three CVaR estimation strategies embedded in $\epsilon$-greedy strategies that are tested are the sample averaging method \eqref{eq:TCEestimsample},  the weighted empirical estimation \eqref{weightedCVaR} and the dual recursive estimation \eqref{recursGen}.

\subsection{Performance metrics}

To assess the performance of the various CVaR estimation methods and corresponding policies, three metrics are considered. At stage $t$ of a given run, the cumulative hit rate $\mathcal{H}_t$ is defined as the percentage of stages in a given run where the action with the smallest current CVaR was sampled:
\begin{equation*}
    \mathcal{H}_{t} \equiv t^{-1} \sum_{u = 1}^{t} \mathds{1}_{  \{ a^*_t = a_t\}}, \quad \text{where } a^*_t \equiv  \underset{i \in \{1,\ldots,K\}  }{\arg\!\min} \, \text{CVaR}_{\alpha}(Y_{t, i}) .
\end{equation*}
A higher cumulative hit rate indicates better identification of the least risky action. A second metric considered is the regret $\mathcal{R}_t$ defined as
\begin{equation*}
    \mathcal{R}_{t} \equiv -\sum_{u = 1}^{t} \left(\min\limits_{i \in \{1,\ldots,K\} } \text{CVaR}_{\alpha}(Y_{u,i}) -  \text{CVaR}_{\alpha}(Y_{u,a_t}) \right).
\end{equation*}
For enhanced interpretability, the average regret $\bar{\mathcal{R}}_t= t^{-1} \mathcal{R}_t$ is reported instead of the cumulative regret in subsequent experiments.
Finally, an empirical (unconditional) CVaR metric relying on the sample averaging method \eqref{eq:TCEestimsample} is defined as
\begin{equation*}
    \mathcal{C}_{t} \equiv \frac{\sum_{u=1}^{t} Y_{u,a_u} \mathds{1}_{ \{ Y_{u,a_u} \geq \hat{q}^{t}_{\alpha}(Y) \} }}{\sum_{u=1}^{t} \mathds{1}_{ \{ Y_{u,a_u} \geq \hat{q}^{t}_{\alpha}(Y) \} }}
\end{equation*}
where
\begin{equation*}
    q^t_\alpha(Y) \equiv \text{inf}\{y \in R: \hat{F}^t_{Y}(y) \geq \alpha \}
\end{equation*}
and $\hat{F}^t_{Y}$ is the empirical distribution obtained with the run's loss sample $\{ Y_{1,a_1},\ldots, Y_{t,a_t}\}$.
The empirical CVaR metric represents a measurement of overall risk faced across all trials, instead of looking at performance trial by trial.
To compute the cumulative hit rate and regret, exact knowledge of the optimal action along with its associated CVaR are required, unlike the empirical CVaR which only depends on observed losses. Note that choosing the action with the smallest CVaR on each stage $t$ does not necessarily corresponds to the sequence of actions $\{a_t\}^T_{t=1}$ leading to the smallest empirical distribution CVaR across all stages. The latter could be calculated based on a dynamic program based on knowledge of loss distributions on each stage, see for instance \cite{godin2016minimizing}, but this complex calculation is not pursued here; the main focus is stage by stage performance, and therefore the metric $\mathcal{C}_{t}$ is merely used as an informal performance indicator rather than the objective function to be optimized.

%%%%%%%%%%%%%%%%%%%%%%%%%%%%%%%%%%%%%%%%%%%%%%%%%%%%%%%%%%%%%%%%%%%%%%%%

\subsection{Simulation experiment}
\label{se:slowly}
%\subsubsection{Settings}

In the simulation experiment all arms produce losses that are normally distributed, with slowly varying parameters. For each arm, the expected value of the loss distribution is randomly generated at the onset of a run based on a uniform distribution, and it remains fixed for the entire run duration. Conversely, the loss distribution standard deviation is also randomly generated at the run onset from a uniform distribution, but it varies progressively on each time step based on an exponential auto-regressive model.
In any given run, such dynamics are summarized by the following equations:
\begin{eqnarray*}
     Y_{t,i} &\sim& \mathcal{N}(\mu_{i} ,  \sigma^2_{t,i}), \quad t=1,\ldots,T,
     \\  \sigma_{t,i}^2 &=& \sigma_{t-1,i}^2 \exp (\epsilon_{t,i}), \quad t=1,\ldots,T,
\end{eqnarray*}
with random shocks $\{ \epsilon_{t,i} \}_{t=1}^{T}$, $i=1,\ldots,8$ forming normal i.i.d. sequences driving the evolution of loss distribution variances, thereby generating non-stationarity.
Distributions considered for the generation of random variables $\epsilon_{t,i}$, $t=1,\ldots,T$, $\mu_{i}$ and $\sigma^2_{0,i}$ are exhibited in Table \ref{TablSpecSimul1}. \textit{Unif}(a,b) denotes a uniform distribution on the $[a,b]$ interval. 

\clearpage

\begin{table}[h]
\caption{Loss distribution parameters specification}
\label{TablSpecSimul1}
\centering
\small
\begin{tabular}{cccc}
\hline
{\bf arm 1} & {\bf arm 2} & {\bf arm 3} & {\bf arm 4}  \\
\hline\\
\vspace{0.05 cm}
%$\mathcal{N}(12.48 , 2.25^2)$  & $\mathcal{N}(13 , 1.5^2)$  & $\mathcal{N}(12.135 , 3.08^2)$  & $\mathcal{N}(15 , 2.4689^2)$ \\
%CVaR  = 9.596  &  CVaR = 11.557 & CVaR = 8.1878 & CVaR = 11.83598 \\
$\mu_1  \sim Unif(0, 2)$ & $\mu_2  \sim Unif(0, 2)$ & $\mu_3  \sim Unif(0, 2)$ & $\mu_4   \sim Unif(0, 2)$ \\
$\sigma_{0,1}  \sim Unif(1, 2)$ & $\sigma_{0,2} \sim Unif(1, 2)$ & $\sigma_{0,3}   \sim Unif(1, 2)$ & $\sigma_{0,4}  \sim Unif(1, 2)$ \\
$\epsilon_{t,1}\sim\mathcal{N}(0, 0.08870^2)$ & $\epsilon_{t,2}\sim\mathcal{N}(0, 0.08871^2)$ &$\epsilon_{t,3}\sim\mathcal{N}(0, 0.08872^2)$ &$\epsilon_{t,4}\sim\mathcal{N}(0, 0.08873^2)$ 
\vspace{0.1 cm} \\
\hline
{\bf arm 5} & {\bf arm 6} & {\bf arm 7} & {\bf arm 8}  \\
\hline\\
\vspace{0.05 cm}
%$\mathcal{N}(16 , 4.89289^2)$ & $\mathcal{N}(20 , 6^2)$ & $\mathcal{N}(18 , 5.5^2)$ & $\mathcal{N}(15 , 4.5^2)$ \\
%CVaR = 9.729 &  CVaR = 12.31069 & CVaR = 10.95147 & CVaR = 9.233018 \\
$\mu_5  \sim Unif(0, 2)$ & $\mu_6  \sim Unif(0, 2)$ & $\mu_7  \sim Unif(0, 2)$ & $\mu_8   \sim Unif(0, 2)$ \\
$\sigma_{0,5}  \sim Unif(1, 2)$ & $\sigma_{0,6} \sim Unif(1, 2)$ & $\sigma_{0,7}   \sim Unif(1, 2)$ & $\sigma_{0,8}  \sim Unif(1, 2)$ \\
$\epsilon_{t,5}\sim\mathcal {N}(0, 0.08874^2)$ & $\epsilon_{t,6}\sim\mathcal{N}(0, 0.08874^2)$ &$\epsilon_{t,7}\sim\mathcal{N}(0, 0.08872^2)$ &$\epsilon_{t,8}\sim\mathcal{N}(0, 0.08873^2)$ 
\end{tabular}
\end{table}

\begin{figure}[h]
\centering
\caption{Evolution of the loss distribution CVaR$_{0.9}$ for all arms in a simulated run.}
\label{CVaRevolCase1}
\begin{minipage}{.9\linewidth}
\includegraphics[scale = 0.43]{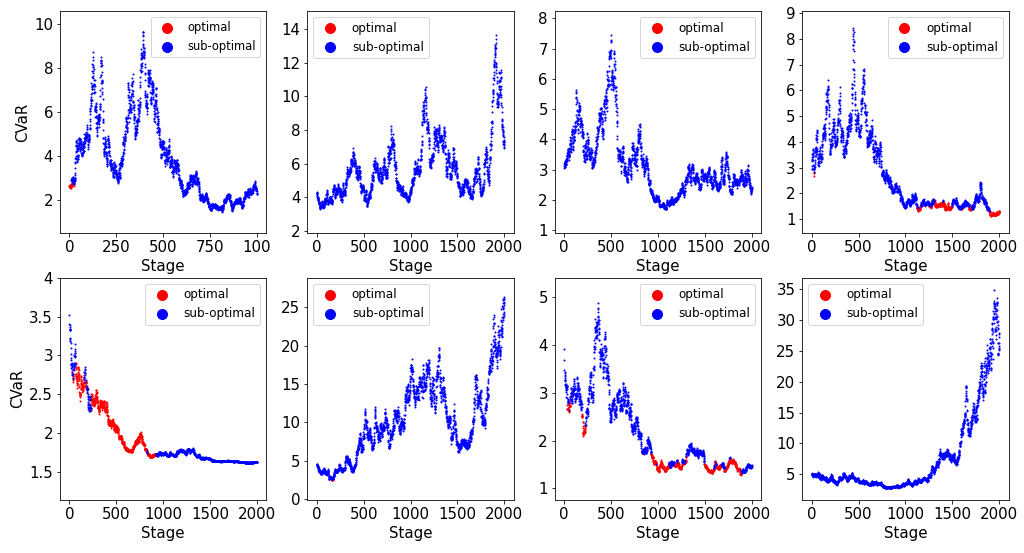} 
%\rule{\linewidth}{10em}
\footnotesize
\emph{Notes:} Each panel exhibits the evolution of the loss distribution CVaR$_{0.9}$ for one of the eight arms over the 2,000 stages in a single simulated run. The red color denotes the arm that is optimal (i.e. having the smallest CVaR$_{0.9}$) on a given stage.
\end{minipage}

%{\footnotesize \\ Each panel exhibits the evolution of the CVaR$_{0.9}$ for one of the eight arms over the 10,000 stages in a single simulated run. The red color is used to denote the arm that is optimal (i.e. having the smallest CVaR) on a given stage.}

%\clearpage
\end{figure}

Simulation parameters from Table \ref{TablSpecSimul1} are chosen such that the loss distributions for all arms are sufficiently close to each other to allow the optimal arm to change within a run instead of constantly having a single arm dominating the others throughout entire runs. 
The evolution of the CVaR for each of the eight arms across the $10,\!000$ stages within a single simulated run is illustrated in Figure \ref{CVaRevolCase1}. Recall that the CVaR$_\alpha$ of a normally distributed random variable $Y$ with mean $\mu$ and standard deviation $\sigma$ is
\begin{equation*}
    \text{CVaR}_\alpha(Y) = \mu + \sigma \dfrac{\phi\left(\Phi^{-1}(\alpha) \right)}{1 - \alpha}
\end{equation*}
with $\Phi$ and $\phi$ being respectively the cdf and the density function (pdf) of a standard normal random variable. In Figure \ref{CVaRevolCase1}, each panel corresponds to an arm, with the red color representing the arm that is optimal on a given stage. The optimal arm indeed varies throughout the run.
%It shows that over the run the variance of the arms may get pretty large/small which may results in some arms having a CVaR very large/very close to its means. %However we can reduce this effect by either reducing the number of steps, reducing the  variance of our lognormals, or both.

%\rc{Figure 1 is too small (cannot read it). We have to make sure its readable. Furthermore, the comments should be left aligned, and space editing should be applied. (working on it)}

\clearpage

\begin{figure}[h]
%\centering

\includegraphics[scale=.33]{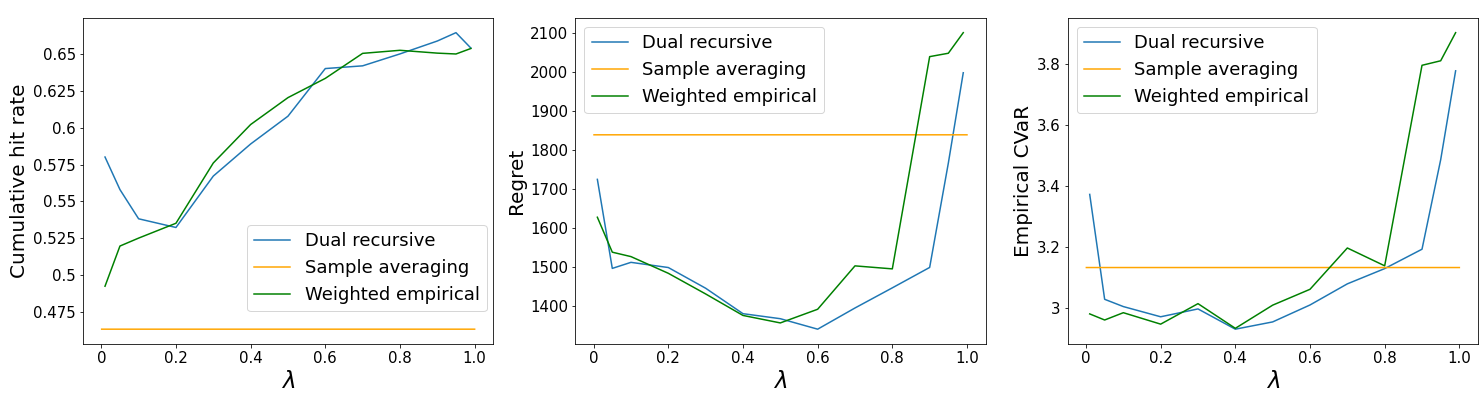}
%\label{fig:rocCurve1}
\caption{Performance versus parameter $\lambda$ (averaged over 1000 runs)}
\label{Fig:PerfStep}

\begin{minipage}{.95\linewidth}
\footnotesize
\emph{Notes:} Time-$T$ performance metrics are reported for various choices of the parameter $\lambda$ for the three estimation methods under the slowly-varying loss distributions experiment detailed in Section \ref{se:slowly}. Parameter $\lambda$ corresponds to an exponential weight decay rate for the weighted empirical estimation, and to a learning rate for the dual recursive estimation. The orange curve corresponds to the sample averaging method \eqref{eq:TCEestimsample}, the green curve to the weighted empirical estimation method \eqref{weightedCVaR}, and the blue curve to the dual recursive approach \eqref{CVaRestimNonstat}. Left panel: cumulative hit rate $\mathcal{H}_t$. Middle panel: average regret $\bar{\mathcal{R}}_t$. Right panel: empirical CVaR $\mathcal{C}_t$.

%\footnotesize Time-$T$ performance metrics are reported for various choices of parameter $\lambda$ for the three estimation methods for the slowly-varying loss distributions experiment detailed in Section \ref{se:slowly}. Parameter $\lambda$ corresponds to an exponential weight decay rate for the weighted estimation, and to a step size for the dual recursive estimation. The orange curve corresponds to the sample averaging method \eqref{eq:TCEestimsample}, the green curve to the weighted estimation method \eqref{weightedCVaR}, and the blue curve to the dual recursive approach \eqref{CVaRestimNonstat}. Left panel: hit rate $\mathcal{H}_t$. Middle panel: average regret $\bar{\mathcal{R}}_t$. Right panel: empirical CVaR $\mathcal{C}_t$.

\end{minipage}
\end{figure}

First, to assess the impact of the choice of the step size $\lambda$ within the dual recursive estimation method or the weight exponential decay rate $\lambda$ in the weighted empirical estimation, a sensitivity analysis is conducted by comparing the time-$T$ (i.e. final) performance metrics for various choices of $\lambda$. Results are provided in Figure \ref{Fig:PerfStep}. Note that the set of generated losses over all arms and stages in a given run (i.e. $\{Y_{t,i}\}^T_{t=1}$, $i=1,\ldots,8$) is shared by the different values of $\lambda$ and the different estimation methods to reduce the impact of randomness when comparing the performance of different methods and hyperparameters.
%In order to reduce the randomness of the experiment as much as possible, for each run, we generate the rewards of all arms across all stages, which will be shared by the three different estimation methods as well as the different stepsizes

%\rc{(Note the Empirical CVaR metric seems to be the metric varying the most, more than 100 runs would be needed to get a better idea of how the Empirical CVaR varies, for the other 2 metrics even though we need more than 100 runs to get better results we already get a good idea of how the different algorithms performs for various stepsizes)}

The orange curve representing the sample averaging method is flat since such method does not depend on any step size related parameter $\lambda$. 
%The weighted empirical estimation method seems less sensitive to the step size than the dual recursive estimation method. 
For both the dual recursive and weighted empirical estimation methods, values around $\lambda=0.5$ seem to provide near-optimal results with respect to the regret and empirical CVaR performance metrics, and good results for the cumulative hit rate. Such value $\lambda=0.5$ is therefore retained for further tests. Compared to values often used in traditional reinforcement learning algorithms, the step size $\lambda=0.5$ could be considered quite large. The good performance associated with such value in the present framework is potentially due to the following: (i) there are not many observations available in the tail of the distribution for each arm, and thus extreme risk quantification requires larger step sizes associated with quicker and stronger updates, and (ii) non-stationarities require the use of larger step sizes than in stationary problems to put larger emphasis on most recent losses.

\begin{figure}[h]

\includegraphics[scale= .33]{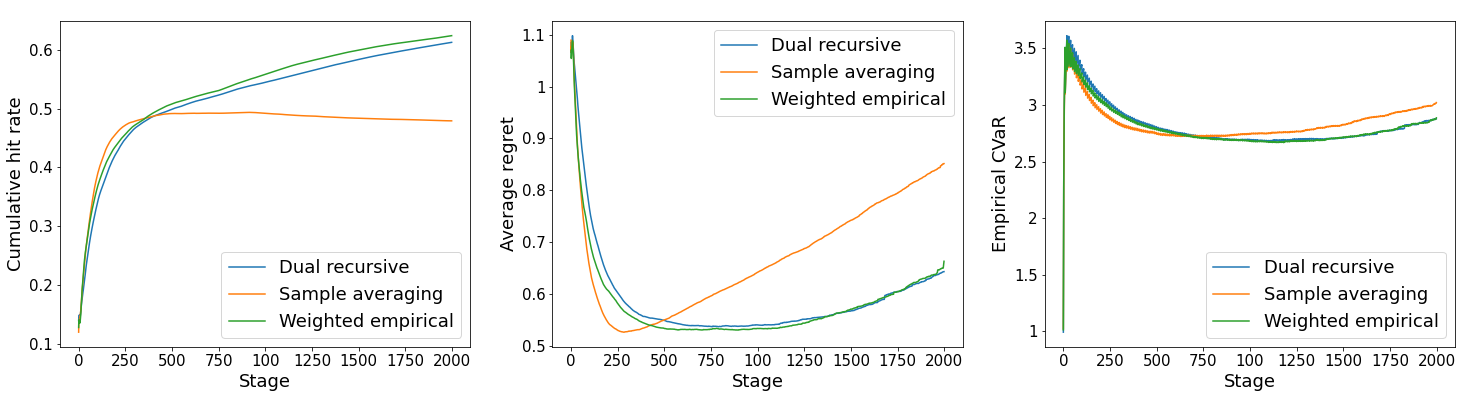}
\label{fig:PerfFixedLambda}
\caption{Evolution of performance metrics over time stages $t$ with $\lambda = 0.5$ (averaged over 1000 runs)}

\begin{minipage}{.95\linewidth}

\footnotesize
\emph{Notes:} Time-$t$ performance metrics are reported over the various stages $t$ for the three estimation methods under the slowly-varying loss distributions experiment detailed in Section \ref{se:slowly}. The orange curve corresponds to the sample averaging method \eqref{eq:TCEestimsample}, the green curve to the weighted empirical estimation method \eqref{weightedCVaR}, and the blue curve to the dual recursive approach \eqref{CVaRestimNonstat}. Left panel: cumulative hit rate $\mathcal{H}_t$. Middle panel: average regret $\bar{\mathcal{R}}_t$. Right panel: empirical CVaR $\mathcal{C}_t$.

%\footnotesize Time-$t$ performance metrics are reported for over the various stages $t$ for the three estimation methods for the slowly-varying loss distributions experiment detailed in Section \ref{se:slowly}. The orange curve corresponds to the sample averaging method \eqref{eq:TCEestimsample}, the green curve to the weighted estimation method \eqref{weightedCVaR}, and the blue curve to the dual recursive approach \eqref{CVaRestimNonstat}. Left panel: hit rate $\mathcal{H}_t$. Middle panel: average regret $\bar{\mathcal{R}}_t$. Right panel: empirical CVaR $\mathcal{C}_t$.

\end{minipage}

\end{figure}

The performance of the three algorithms with a step size of $\lambda=0.5$ is now compared. The various performance metrics are reported in Figure \ref{fig:PerfFixedLambda} for all stages $t$ and all three estimation algorithms.

The weighted averaging (green curve) and dual recursive (blue curve) algorithms clearly outperform the sample averaging (orange curve) in the long run, which is not surprising given the presence of non-stationary losses. Moreover, the performance of the dual recursive approach and the weighted averaging approach is quite similar.
%can be deemed small due to the high sensitivity of the method due to the impact of the step size selection previously shown in Figure \ref{Fig:PerfStep}; different choices could have led the weighted empirical estimation method to outperform the dual recursive approach. 
Nevertheless, the ability to use the dual recursion formula on-line without needing to store all observed losses can compel a user to use this method for problems with a large number of stages requiring a quick implementation. %Thus, both the weighted averaging and dual recursive method possess different advantages.

\section{Conclusion}
\label{se:Conclusion}

The present paper is the first to introduce action selection methods in the context of non-stationary risk averse multi-armed bandits problems. The objective function considered is the CVaR. Two approaches for its estimation on the various arms are proposed in the context of non-stationarity, one relying on a weighted empirical distribution of losses, and another based on the dual representation of the CVaR involving a recursive update formula. Such recursive formula is convenient if the multi-armed bandits problem is applied in an on-line setting, as it can be paralellized and it does not require storing the history of incurred losses for all arms. Conversely, the approach based on the weighted loss distributions possesses the advantage of not depending on arbitrary initial estimates. The proposed methods are benchmarked against a sample averaging estimator for the CVaR disregarding the potential non-stationary nature of loss distributions within an exploratory simulation experiments with slowly evolving loss distributions. The experiment reveals that the two proposed methods clearly outperform the naive sample averaging benchmark when losses are non-stationary. Moreover, the optimal step size $\lambda$ in the dual recursive estimation update formula (or alternatively the optimal weight decay parameter $\lambda$ in the weighted empirical estimation) is quite large, much above values that are often considered in a stationary case where the objective function is the expectation. Finally, for a suitably chosen parameter $\lambda$, the two non-stationary estimation methods seem to produce similar performance, without one clearly dominating the other.

As future research, it could be interesting to see if extreme value theory \citep[see for instance][]{mcneil2015quantitative} could be used in the context of non-stationary risk averse bandits when confidence levels close to one are considered. Such an approach is applied in the context of stationary bandits in \cite{troop2019risk}, but its extension to the non-stationary case might be non-trivial. Time varying thresholds in the peaks-over-threshold approach as in \cite{kysely2010estimating} could be contemplated as method to tackle this problem. Furthermore, a second interesting research avenue would be the derivation of concentration bounds on CVaR estimates based on assumptions limiting the magnitude of loss distributions fluctuations.

%%%%%%%%%%%%%%%%%%%%%%%%%%%%%%%%%%%%%%%%%%%%%%%%%%%%%%%%%%%%%%%%%%%%%%%%%%%%%%%%%%
%%%%%%%%%%%%%%%%%%%%%%%%%%%%%%%%%%%%%%%%%%%%%%%%%%%%%%%%%%%%%%%%%%%%%%%%%%%%%%%%%%
%%%%%%%%%%%%%%%%%%%%%%%%%%%% BIBLIOGRAPHY %%%%%%%%%%%%%%%%%%%%%%%%%%%%%%%%%%%
%%%%%%%%%%%%%%%%%%%%%%%%%%%%%%%%%%%%%%%%%%%%%%%%%%%%%%%%%%%%%%%%%%%%%%%%%%%%%%%%%%
%%%%%%%%%%%%%%%%%%%%%%%%%%%%%%%%%%%%%%%%%%%%%%%%%%%%%%%%%%%%%%%%%%%%%%%%%%%%%%%%%%

%\newpage
\bibliographystyle{apalike} \bibliography{Bib}

 %%% Uncomment this line and comment out the ``thebibliography'' section below to use the external .bib file (using bibtex) .

%%% Uncomment this section and comment out the \bibliography{references} line above to use inline references.
% \begin{thebibliography}{1}

% 	\bibitem{kour2014real}
% 	George Kour and Raid Saabne.
% 	\newblock Real-time segmentation of on-line handwritten arabic script.
% 	\newblock In {\em Frontiers in Handwriting Recognition (ICFHR), 2014 14th
% 			International Conference on}, pages 417--422. IEEE, 2014.

% 	\bibitem{kour2014fast}
% 	George Kour and Raid Saabne.
% 	\newblock Fast classification of handwritten on-line arabic characters.
% 	\newblock In {\em Soft Computing and Pattern Recognition (SoCPaR), 2014 6th
% 			International Conference of}, pages 312--318. IEEE, 2014.

% 	\bibitem{hadash2018estimate}
% 	Guy Hadash, Einat Kermany, Boaz Carmeli, Ofer Lavi, George Kour, and Alon
% 	Jacovi.
% 	\newblock Estimate and replace: A novel approach to integrating deep neural
% 	networks with existing applications.
% 	\newblock {\em arXiv preprint arXiv:1804.09028}, 2018.

% \end{thebibliography}

\end{document}